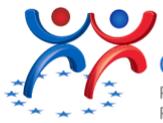 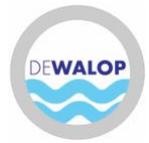

# In-pipe Robotic System for Pipe-joint Rehabilitation in Fresh Water Pipes

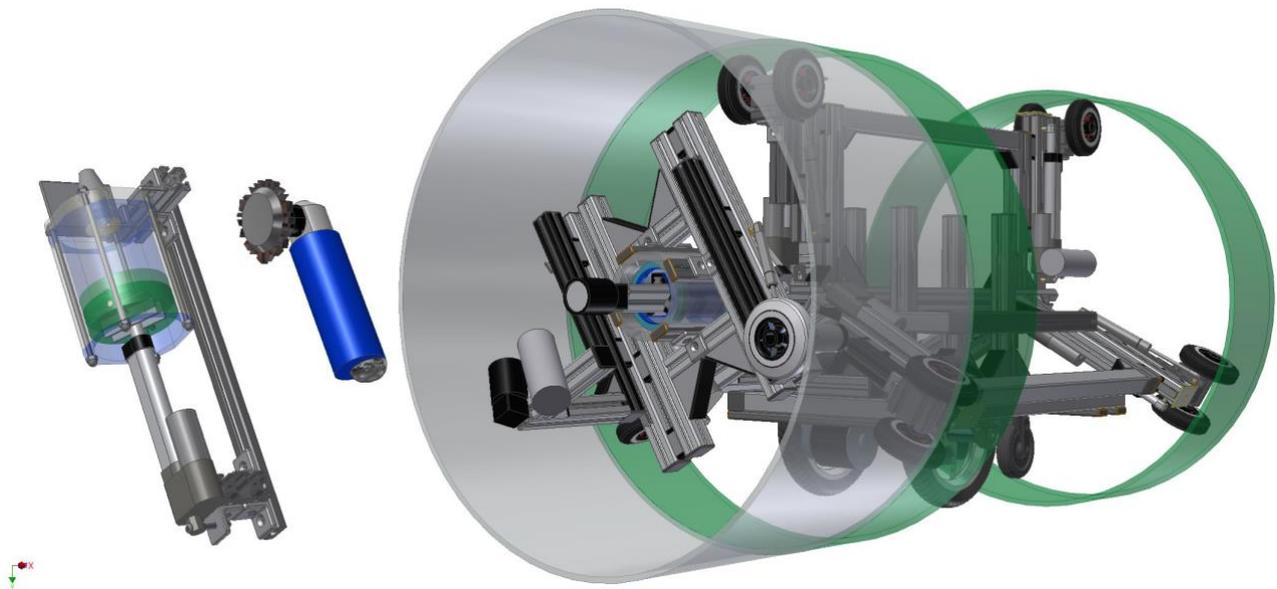

Luis A. Mateos and Markus Vincze

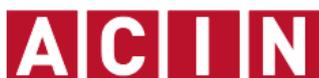
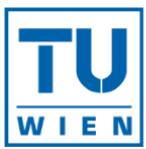

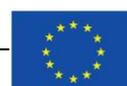

## 1. Introduction

A problem spotted in the city of Vienna and Bratislava is that the quantity of water sent from the municipal systems mismatch with the quantity of water received by the consumers (houses, buildings and companies). The mismatch in the water readings is considerably high, in the range of 15 to 20% [Wasserwerke, 2009].

The Vienna Waterworks looked deeper into the problem and found that the main water loss points are caused by 100km of aged cast-iron pipelines with lead-joint sockets. The sockets were caulked up to the 1920's with a hemp pack and a lead ring. The swollen hemp pack ensured the sealing and the lead ring stabilized the hemp in the socket, as shown in Fig. 1. Over the years the hemp pack decomposed, and the lead ring was displaced by pipe movements leading eventually to leakage [Kottmann, 1997].

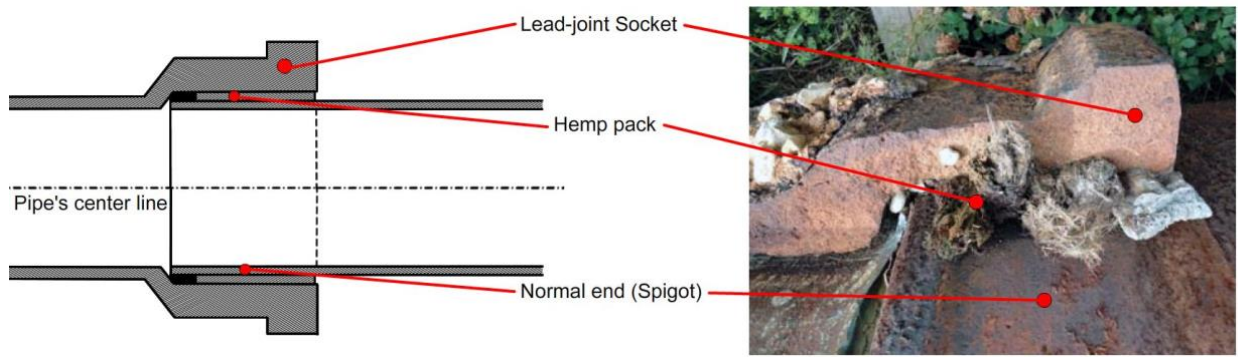

Figure 1: Pipe joint diagram (left). Lead joint socket with worn out hemp pack and lead ring (right).

From the water loss study, the pipeline segments with leakages may be known. How- ever, these segments are in the order of kilometers, making not feasible pipe replacement due to the high cost involved. As consequence, the preferred solution is pipe rehabilitation, which involves inspection, cleaning and repairing [Amir and Kawamura, 2007].

### 1.2 Human operators inside the pipe

Initial attempts to rehabilitate these types of pipes included operators inside the one-meter diameter pipes, as shown in Fig. 2. Although, this creates a special situation that presents a safety and healthy risk to human operators [S. Yang and Kwon, 2008].

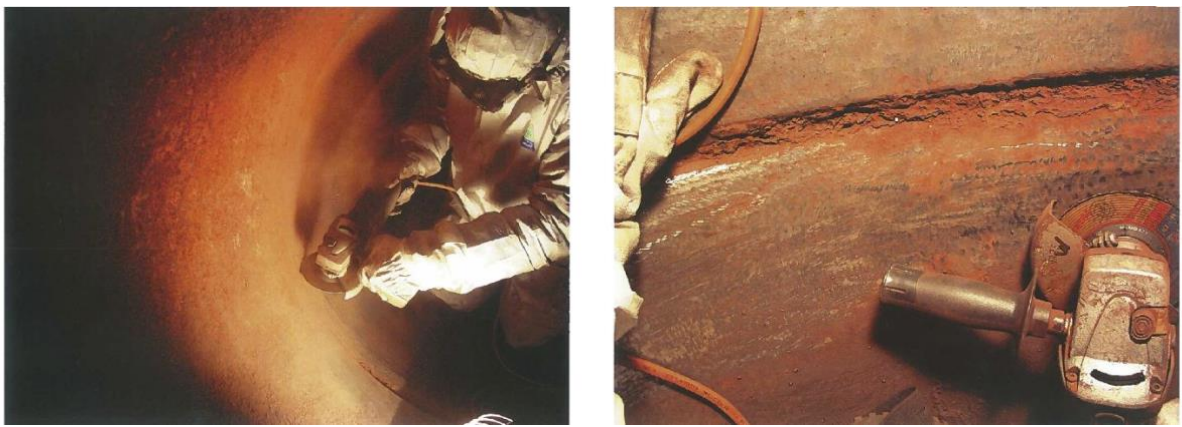

Figure 2: Operator inside a 1000mm diameter cast-iron pipe cleaning the pipe joint.

### 1.3 Failed sealing system

In the city of Vienna, a Silane Modified Polyether (SMP) sealant was applied to rehabilitate detected water-loss pipes. However, the system failed after 2 years in service; the investigation showed that both biological and environmental influences caused the failure [V. Archodoulaki and Werderitsch, 2010].

Other types of solutions, such as sleeve or 'patch' repairs, resin injection and mechanical joint sealing were discarded due to high cost and the need of workers inside the pipe.

## 2. Problem statement

The pipe-joint socket leakages lead to erosion of pipe bedding, imposing stress to the pipe, increasing the pipe break likelihood and adding the risk of pipe burst where pipelines are located under heavy traffic roads [B. Rajani and Kuraoka, 1996].

### 2.1 In-pipe robots

Currently, the applications of robots for the maintenance and rehabilitation of the pipeline utilities are considered as one of the most attractive solutions available. Since, no human operator is required inside the pipe and the robot can integrate multiple sensors, cameras and tools to inspect and repair [Roh et al., 2009].

Nevertheless, for an in-pipe robot to substitute a skilled human operator, pipe re-habilitation requires mechanisms with high degree of mobility, able to move along the pipelines, overcoming obstacles, extreme environments and with high accuracy clean and repair specific areas of the pipe [Li et al., 2007b].

Commercial robotic platforms have several drawbacks, if considering to be used in freshwater pipes. The first downside is their integrated hydraulic system, which includes liquids that are harmful to humans, and therefore, these robots should not operate in freshwater pipes. Additionally, their working pipe diameters are constrained to small-medium pipes, ranging from 200 to 800mm diameter. Moreover, commercial in-pipe cleaning robots commonly include a mobile robot similar to a vehicle with four wheels. Which is not fixed to the center of the pipe when cleaning. Hence, when the robot is cleaning (grinding or milling) with its power tool, the stability of the robot relays only on the friction of the wheels to the pipe surface given by the robot's weight, creating a dangerous scenario. Also, these robots do not include any suspension system to damp vibration forces from the cleaning tool, which may damage or even break the pipe.

### 2.2 In-pipe robots for freshwater pipelines

In order to repair freshwater pipelines, the in-pipe robots, as prerequisite should not contain any liquid or chemical harmful to humans. Consequently, the robotic mechanisms must integrate pneumatic and/or electrical systems.

For the specific case of aged cast-iron pipes supplying water to the municipal pipeline networks, the in-pipe robots must be able to repair in non-standard pipe sizes with diameters 800 to 1200mm. To do so, the robot must be set perfectly to the center of the pipe, so the vibrations generated by the power tool when cleaning are damped by its integrated suspension system and not by the pipe. Since, cast-iron is a fragile and easy to break material.

## 3. DeWaLoP - Developing Water Loss Prevention

DeWaLoP is a project for cross-border cooperation between Austria (Vienna) and Slovakia (Bratislava) 2007 - 2013, within the region of CENTROPE. The aim of the resulting partnership is the solution of environmental problems in connection with the activities in the supply of water to the public in the two cities involved.

The main requirement of the project is the design and development of an in-pipe robotic platform, able to crawl in fresh-water pipes of about one meter in diameter, inspect, clean and apply a sealant to repair the pipe.

## 4. Contributions

The main contribution of the PhD thesis is the design and development of an in-pipe robotic system able to crawl 100m inside pipes with diameters ranging from 800 to 1200mm. Inspect details inside the pipe with its integrated multi-camera system. Fix its position to the center of the pipe to prevent damages to the pipe while cleaning and sealing it.

### 4.1 In-pipe robot

In contrast to the in-pipe robotic mechanisms state-of-the-art, DeWaLoP robot is able to fix itself centered and stably in a specific location inside the pipe using its wheeled-legs and suspension system [Mateos and Vincze, 2011e] [Mateos et al., 2012]. And independently from the main body of the robot, the tool system can be flexibly adjusted in a cylindrical 3D space, covering the entire pipe-joint area. Able to overcome axial-pipe displacements of 100mm, while reaching to the surface of pipes with diameters in the range of 800mm to 1200mm [Mateos and Vincze, 2012a] [Mateos et al., 2013].

### 4.1.1 DeWaLoP robot modules

The DeWaLoP in-pipe robot consists of five modules, four of them are integrated in the in-pipe robot: the mobile robot, maintenance unit, tool and vision systems. While the control station is located outside the pipe, see Fig. 3.

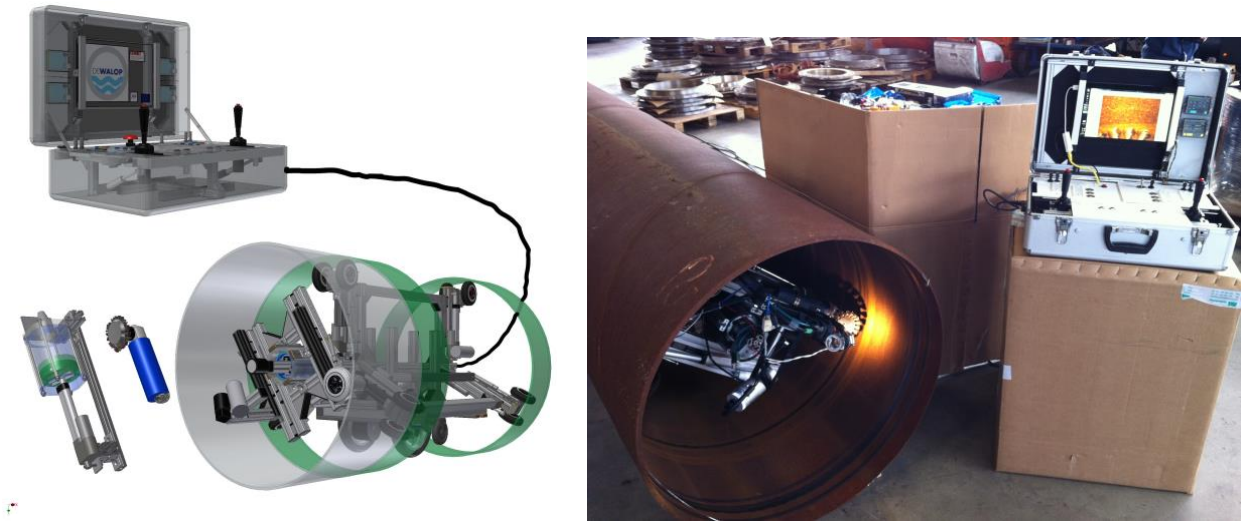

Figure 3: DeWaLoP robotic system 3D model (left). Control station (remote control) outside the pipe. DeWaLoP in-pipe robot modules: 1) Mobile robot. 2) Maintenance unit. 3) Tool system. 4) Camera system. Robot tested at Vienna Waterworks (right).

**Control station**

The control station monitors and controls all the components of the in-pipe robot. The controller includes a slate computer to monitor and display the video images from the robot's Ethernet cameras. And several 8 bits micro-controllers with Ethernet capabilities to send commands to the in-pipe robot and to receive data from the robot's sensors and actuators [Mateos et al., 2011].

**Mobile robot**

The mobile platform is able to move inside the pipes, carrying on board the in-pipe robot modules: maintenance unit, tool and camera systems. Also, electronic and mechanical components of the robot, such as: motor drivers, power supplies, etc. It uses a differential wheel drive, which enables the robot to promptly adjust its position to remain in the middle of the pipe while moving [Mateos and Vincze, 2011b].

**Maintenance unit**

The maintenance unit structure consists of six wheeled legs; distributed in pairs of three, on each side, separated by an angle of 120° supporting the structure along the center of the pipe. The maintenance unit combines a wheel-drive-system with a wall-press-system, enabling the robot to operate in pipe diameters varying from 800mm to 1200mm. Moreover, the maintenance unit together with the mobile robot forms a monolithic multi-module robot, which can be easily mounted/dismounted without the need of screws [Mateos and Vincze, 2011d].

The wheeled legs are able to extend or compress with a Dynamical Independent Suspension System (DISS) [Mateos and Vincze, 2011e]. When extending its wheeled legs, the legs create a rigid centered structure inside the pipe. While, when compressing its wheeled legs, the wheels become active and the maintenance unit is able to move along the pipe by the mobile robot.

**Tool and vision system**

The tool system enables the rehabilitation of the pipe by cleaning and sealing it [Mateos and Vincze, 2012a] [Mateos and Vincze, 2013b]. The in-pipe robot vision system includes four cameras, in order to navigate, detect defects and repair specific areas in the pipes [Mateos and Vincze, 2011a].

**5 Experimental results**

The pipe-joint rehabilitation process includes: inspection, cleaning (see Fig. 4.), sealant preparation and sealant application (see Fig. 5).

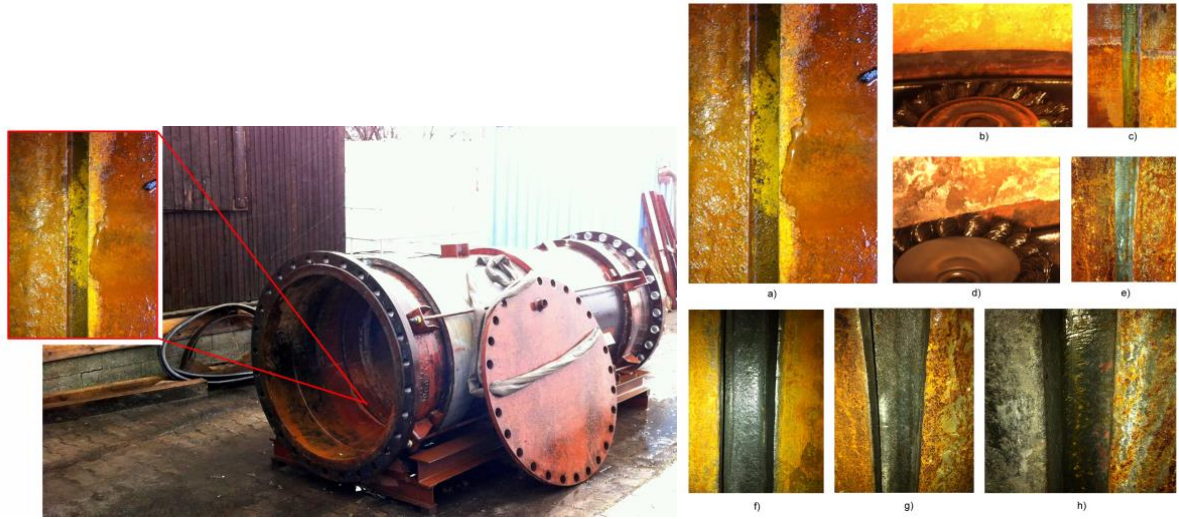

Figure 4. Results from pipe-joint cleaning. a) Corroded pipe-joint inside the pressure chamber. b) Cleaning process with straight knotted-wire-brushes disk. c) Preliminary cleaning results, revealing medium corrosion levels. d) Cleaning process with tapered knotted wire-brushes disk. e) Preliminary cleaning results reveal cleaner pipe-joint areas. f) Around 95 to 99% of corrosion removed. g) Around 90% of corrosion removed, still small areas with corrosion can be seen. h) Around 85% of corrosion removed, revealing corroded points in groove areas. (The sealant requires at least 80% corrosion removal in the area of application).

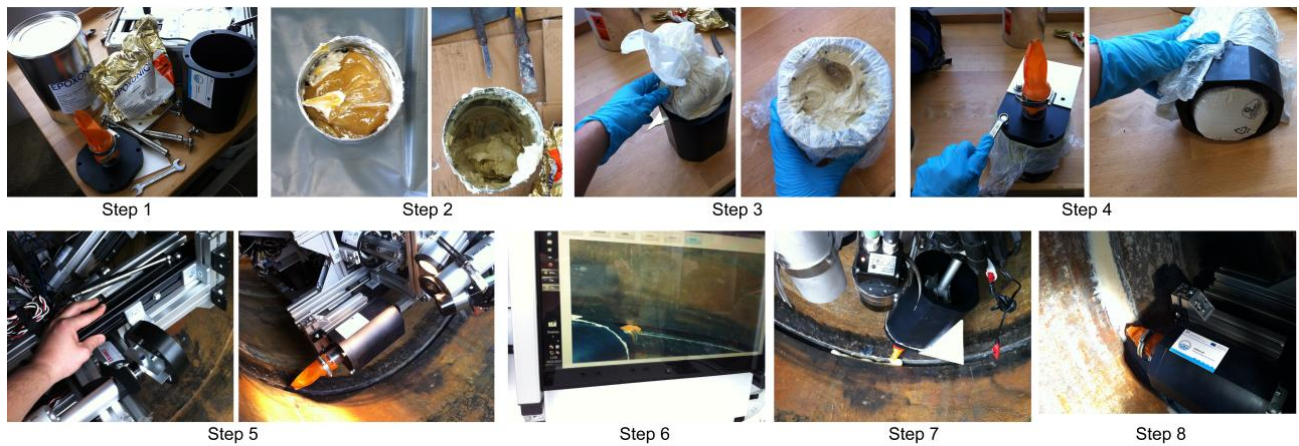

Figure 5. Sealant preparation and robotic application. Step 1. The cartridge from the injection system is opened and ready to be charged with the sealant material. Step 2. The high viscosity epoxies are mixed, resulting in a homogeneous material. Step 3. The sealant is into the injection system cartridge. Step 4. The cartridge is closed with its front cover, while on the opposite end it is still open to be covered and push by the inner container cover. Step 5. The cartridge is then mounted on the injection system. The linear actuator is ready to push the material. The robot is moved inside the pipe and stopped over a pipe- joint. The robot adapted its maintenance unit configuration, from compressed to extended mode, forming a centered and rigid structure inside the pipe. Once the extending finished, the sealing mechanism is enabled and selected. The arm with the drive-wheel is extended first until it makes proper contact to the pipe surface, by compressing partially its spring. Then, the arm with the sealing tool is moved to the desired position in cylindrical 3D space. Step 6. The arm with the sealing tool is moved to the pipe-joint, until the nozzle is located inside the pipe-joint. Step 7. The sealing system starts to inject the sealant progressively, while moving around the inner pipe circumference. The linear actuator slowly pushes the material with 1000N force at a speed of 6mm per second. Step 8. The use of a spatula tool gives a smooth finish to the sealant material.

**For more information about the positive results from the Thesis, please visit:**

http://www.dewalop.eu/index.php/press-corner

http://www.wien.gv.at/rk/msg/2014/04/02001.html